\documentclass[10pt, a4paper]{article}

\usepackage{arxiv}

\usepackage[utf8]{inputenc} % allow utf-8 input
\usepackage[T1]{fontenc}    % use 8-bit T1 fonts
\usepackage{hyperref}       % hyperlinks
\usepackage{url}            % simple URL typesetting
\usepackage{booktabs}       % professional-quality tables
\usepackage{amsfonts}       % blackboard math symbols
\usepackage{nicefrac}       % compact symbols for 1/2, etc.
\usepackage{microtype}      % microtypography
\usepackage{lipsum}		% Can be removed after putting your text content
\usepackage{graphicx}
\usepackage{natbib}
\usepackage{doi}
\usepackage{multirow,tabularx}
\usepackage{caption}
\usepackage{subcaption}
\usepackage[dvipsnames]{xcolor}

\title{Swiss German Speech to Text evaluation}

\date{} 					% Or removing it
\author{
Yanick Schraner, Christian Scheller, Michel Plüss, Manfred Vogel \\
University of Applied Sciences and Arts Northwestern Switzerland
}

\renewcommand{\shorttitle}{Swiss German Speech to Text system evaluation}

\hypersetup{
pdftitle={Swiss German Speech to Text evaluation},
pdfsubject={q-bio.NC, q-bio.QM},
pdfauthor={Yanick Schraner, Christian Scheller, Michel Plüss, Manfred Vogel},
pdfkeywords={First keyword, Second keyword, More},
}

\begin{document}

\maketitle

\begin{abstract}
	
We present an in-depth evaluation of four commercially available Speech-to-Text (STT) systems for Swiss German.
The systems are anonymized and referred to as system a, b, c and d in this report.
We compare the four systems to our STT models, referred to as FHNW in the following, and provide details on how we trained our model.
To evaluate the models, we use two STT datasets from different domains.
The Swiss Parliament Corpus (SPC) test set and the STT4SG-350 corpus, which contains texts from the news sector with an even distribution across seven dialect regions.
We provide a detailed error analysis to detect the strengths and weaknesses of the different systems.
On both datasets, our model achieves the best results for both, the WER (word error rate) and the BLEU (bilingual evaluation understudy) scores.
On the SPC test set, we obtain a BLEU score of $0.607$, whereas the best commercial system reaches a BLEU score of $0.509$.
On the STT4SG-350 test set, we obtain a BLEU score of $0.722$, while the best commercial system achieves a BLEU score of $0.568$.
However, we would like to point out that this analysis is somewhat limited by the domain-specific idiosyncrasies of the selected texts of the two test sets.

\end{abstract}


\section{Introduction}
Swiss German is a family of German dialects spoken by around five million people in Switzerland. 
It differs from Standard German regarding phonetics, vocabulary, morphology, and syntax and is primarily a spoken language.
While it is also used in writing, particularly in informal text messages, it lacks a standardized orthography.
This leads to difficulties for automated text processing due to spelling ambiguities and the huge vocabulary size.
Therefore, it is often preferable to work with Standard German text, for which automated processing tools exist in abundance.

One way to tackle Swiss German ASR is an end-to-end Swiss German speech to Standard German text approach.
This can be viewed as a speech translation (ST) task with similar source and target languages. In our analysis we therefore report the word error rate (WER) and the BLEU score (\cite{papineni2002}), as this is common in ST tasks.
To our knowledge, four public datasets are available with Swiss German audio and Standard German transcripts:
\begin{itemize}
\setlength\itemsep{0em}
    \item The Radio Rottu Oberwallis dataset (\cite{garner2014}) consists of 7 hours of Wallis dialect audio with clean labels, of which 2 hours are transcribed in Standard German.
    \item The SwissDial dataset (\cite{dogan2021}) consists of 26 hours of audio from 8 different dialects with clean labels.
    \item The Swiss Parliaments Corpus (SPC) (\cite{pluess2021a}) consists of 293 hours of mostly Bern dialect audio with noisy labels.
    \item The SDS-200 dataset (\cite{pluess2022}) consists of 189 hours of audio from speakers from all parts of German-speaking Switzerland with clean labels.
\end{itemize}
As can be seen in this list, before 2021, there was almost no publicly available training data for Swiss German ASR, which made the task very hard. With the new data, training end-to-end models is now feasible.

The main contribution of this research is a comparison between four commercially available ASR systems and our ASR models.
We evaluate the systems on corpus, speaker, and dialect levels.
In addition, we analyze the performance on sentences containing named entities.
%\textcolor{red}{For our ASR system, we perform an out-of-vocabulary analysis}.
We use two different test sets.
The test set of the STT4SG-350 corpus (\cite{pluess2023}) consists of 35 hours of read news and parliament minutes sentences in 7 different dialects.
The SPC test set contains 6 hours of parliament speeches in the Bern dialect.

The remainder of this paper is structured as follows: The characteristics of the two test corpora are detailed in section \ref{sec:data}.
We describe our own models in section \ref{sec:model} and the actual evaluation is done in section \ref{sec:evaluation}.

\section{Evaluation Data}
\label{sec:data}
\subsection{The STT4SG-350 test data}
\begin{figure}
    \centering
    \begin{subfigure}[b]{0.3\textwidth}
        \centering
        \includegraphics[width=\textwidth]{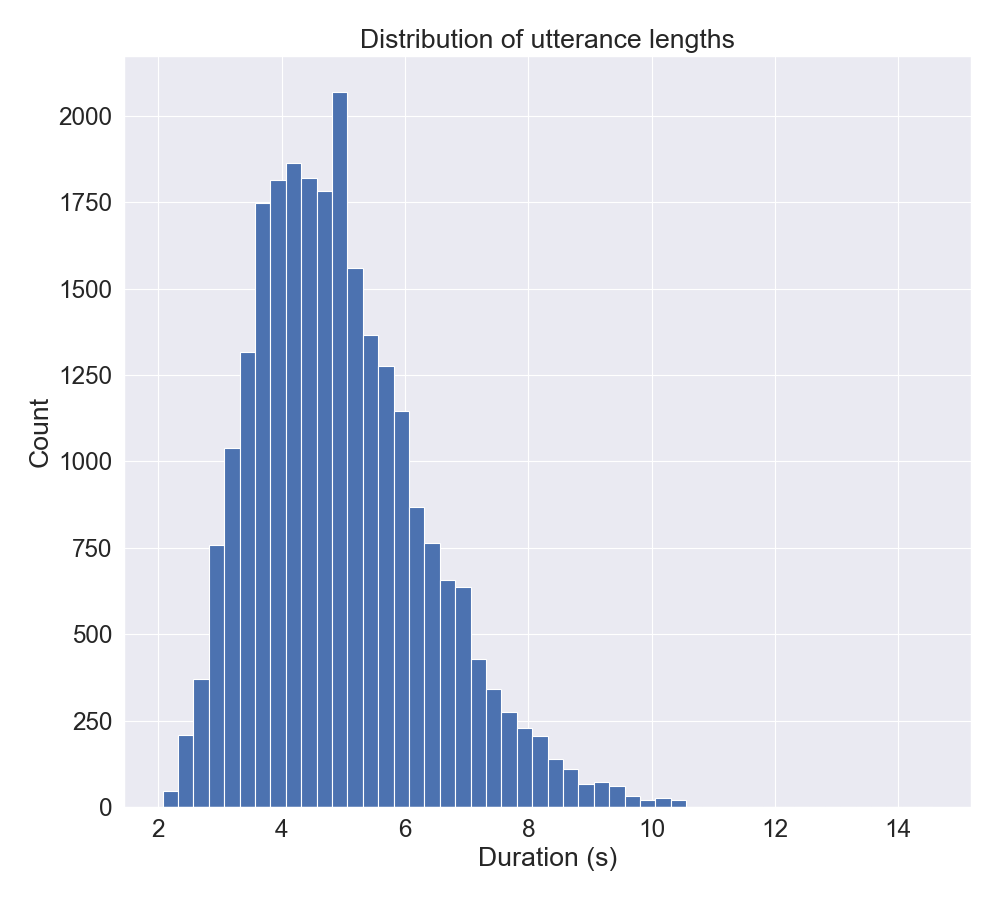}
        \caption{Distribution of utterance lengths in the test dataset.}
        \label{fig:utterance_dist}
    \end{subfigure}
    \hfill
    \begin{subfigure}[b]{0.3\textwidth}
        \centering
	    \includegraphics[width=\textwidth]{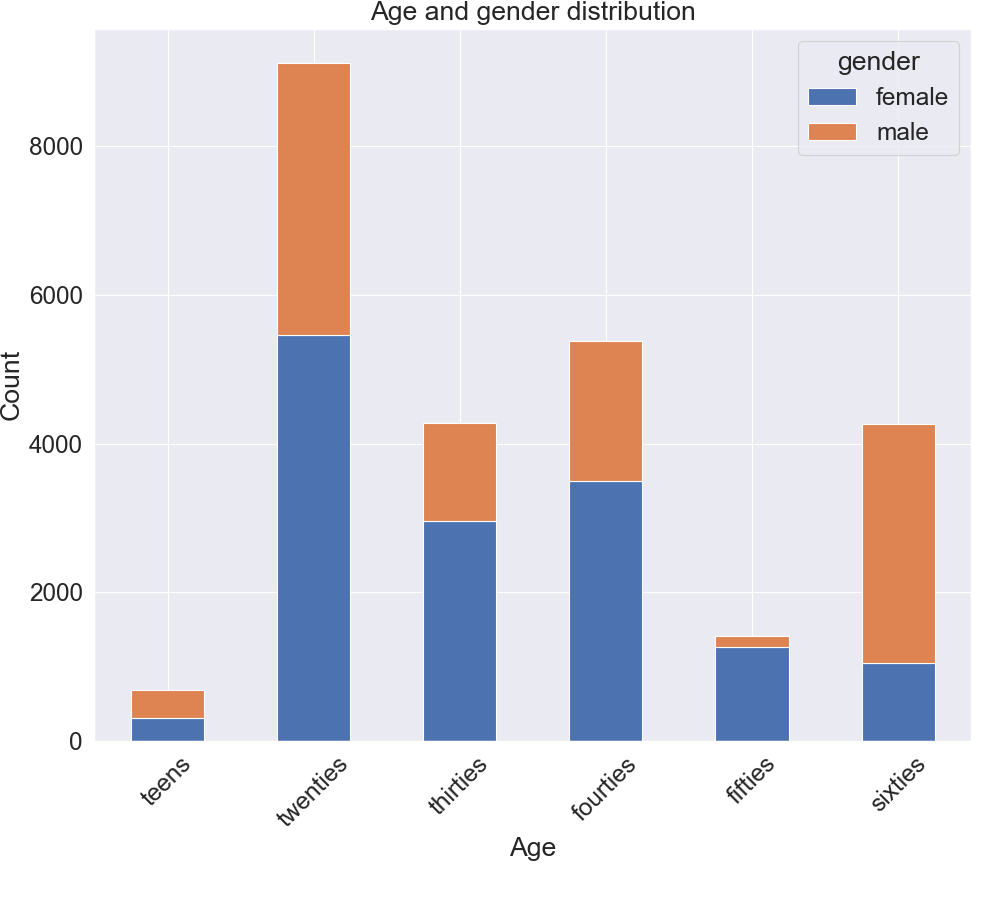}
	    \caption{Number of utterances per speakers' age group and gender.}
	    \label{fig:age}
    \end{subfigure}
    \hfill
    \begin{subfigure}[b]{0.3\textwidth}
        \centering
	    \includegraphics[width=\textwidth]{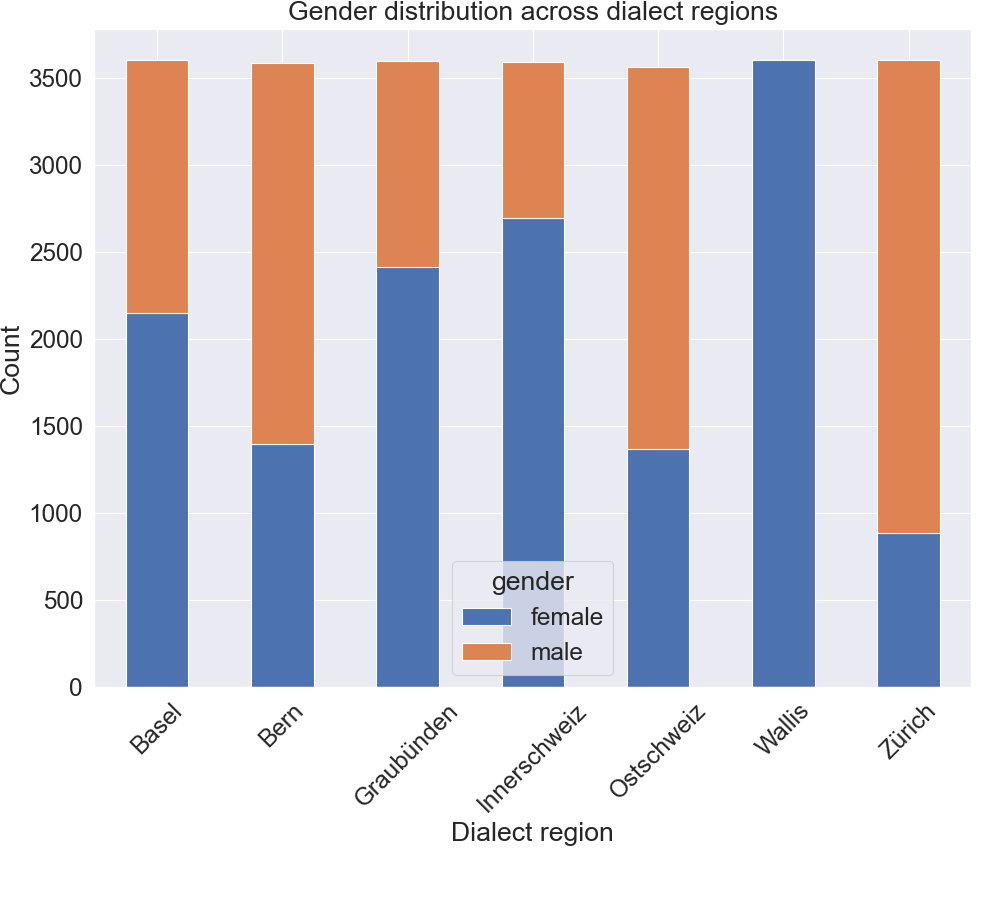}
	    \caption{Gender distribution of dialect regions.}
	    \label{fig:gender}
    \end{subfigure}
    \label{fig:test_set_data}
    \caption{STT4SG-350 test set characteristics.}
\end{figure}

The STT4SG-350 corpus (\cite{pluess2023}) contains a test set with a total of $25'144$ utterances, hence audio with a total length of 35 hours.
The audio was collected with a webapp very similar to \cite{pluess2022}. The only major difference was the sampling of the sentences.
To ensure an equal vocabulary when comparing the performance on different dialects, the same $3'602$ sentences have been recorded by speakers of the following seven dialect regions in Switzerland, roughly 10 speakers per region: Basel, Bern, Graubünden, Innerschweiz, Ostschweiz, Wallis and Zürich.
70 out of the $25'214$ recordings were corrupt and therefore excluded from the STT4SG-350 test set.
On average, an utterance is $5.0$ seconds long with a standard deviation of $1.4$ seconds. The shortest and longest utterances are $2$ and $14.6$ seconds long, respectively.
In Figure \ref{fig:utterance_dist} we display the utterance length distribution.

Out of 76 speakers, 36 are male and 40 are female.
The age and gender distribution over the recorded utterances is given in Figure \ref{fig:age}.
The dialect region Wallis contains only female speakers because, during the recruitment phase, no male speakers could be recruited from Wallis.

\subsection{SPC Test Corpus}
\begin{figure}[htb]
    \centering
    \includegraphics[width=0.49\textwidth]{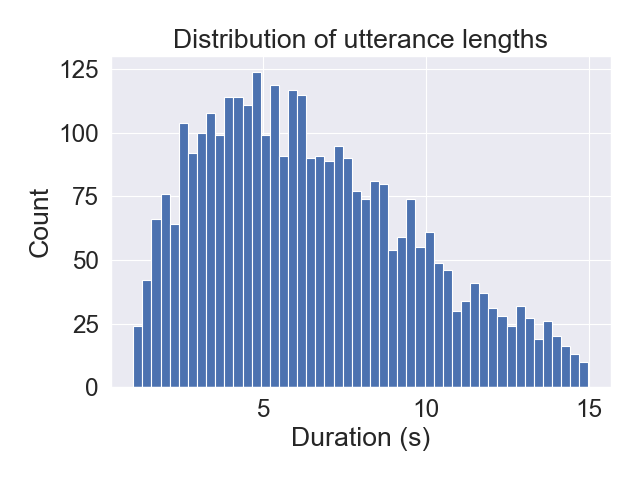}
    \caption{Distribution of utterance lengths in the SPC test set.}
    \label{fig:utterance_dist_spc}
\end{figure}

The test set of the Swiss Parliament Corpus (\cite{pluess2021a}) contains $3'332$ utterances by 26 different speakers. On average, an utterance is 6.5 seconds long with a standard deviation of 3.2 seconds. The shortest and longest utterances are 1 and 15 seconds long, respectively.
In figure \ref{fig:utterance_dist_spc}, we display the utterance length distribution of the SPC test set. In total, we have 6 hours of test data. On average each speaker voiced 128.2 utterances with a standard deviation of 82.8 utterances. The lowest and highest number of voiced utterances per speaker are 2 and 270, respectively.
We do not have metadata like gender, age and dialect of the speakers.
However, the corpus is based on speeches at the Bernese cantonal parliament. Therefore, almost all of the utterances are expected to be in the Bern dialect.
The recordings in the SPC dataset generally have a higher background noise compared to the STT4SG-350 corpus.

\section{Models}
\label{sec:model}
Our model is based on the XLS-R 1B model (\cite{babu2021xlsr}) that was pre-trained on 436K hours of unlabeled speech data covering more than 128 languages. This model is publicly available\footnote{\url{https://github.com/pytorch/fairseq/tree/main/examples/wav2vec/xlsr}}.
Swiss German was not part of the training data. XLS-R Wav2vec models consist of a convolutional feature encoder, followed by a stack of transformer blocks. Details of the architecture configurations can be found in~\cite{babu2021xlsr}. For the finetuning on Swiss German data, we followed the procedure and hyper-parameters described by the authors.
For the finetuning of the XLS-R 1B model we use the following datasets: SDS-200 (\cite{pluess2022}), SPC (\cite{pluess2021a}) and SwissDial (\cite{dogan2021swissdial}).

We use KenLM (\cite{heafield2011kenlm}) to train 4-gram language models. We combined Europarl (\cite{koehn2005europarl}), news-crawl 2019 (\cite{barrault2019findings}), ParlSpeech v2 (\cite{rauh2020parlspeechv2}), and SPC-public and SPC-private train texts to obtain a total of 67 Million German sentences.
The language model is used during decoding with a beam width of 200.

We trained and evaluated an additional model for the STT4SG-350 test set that does not rely on unsupervised training data.
Training such a model allows us to compare a model that heavily uses unsupervised learning on an enormous data set to a simple supervised learning approach.
This model is trained on following datasets: SDS-200, SPC, SwissDial, and Commonvoice German.
We employed Transformer (\cite{vaswani2017attention}) based models implemented in the FAIRSEQ S2T libraryv(\cite{ott2019fairseq,wang2020fairseqs2t}).
These models consist of a two-layer convolutional subsampler followed by a Transformer network with 12 encoder layers and six decoder layers.
We employed eight attention heads for the Transformer network, an embedding dimension size of 512, and a dropout rate of 0.15. We used the default model hyper-parameters and learning rate schedules provided by the library without any task-specific tuning.
This model is denoted as FHNW Transformer in our evaluation.

Instead of a KenLM language model, we train a Transformer-based language model (LM) with 12 decoder layers, 16 attention heads, an embedding dimension of 512, and a fully connected layer with 1024 units.
The LM is trained on the same 67M Standard German sentences as the KenLM model.
We use a beam width of 60 during decoding.

\section{Evaluation}
\label{sec:evaluation}
The evaluations were carried out in May 2022.
We split the evaluation into two subsections, one for each test corpus.
We report the BLEU score and WER on a corpus level.
Additionally, we analyze the influence of named entities on the BLEU score.
We show examples of sentences with low scores in the five tested ASR systems.
For the STT4SG-350 test set, we use the available metadata to report the metrics on a dialect, age, and gender level to show the influence of those variables.
To calculate the WER and BLEU score we normalize the outputs of the various ASR systems to a common vocabulary.

\subsection{STT4SG-350 Test Set}
In Table \ref{tab:results_full} we report the BLEU score and WER of all systems on a corpus and dialect level.
Our models, FHNW XLS-R and FHNW Transformer, are described in Section \ref{sec:model}.
Systems b, c and d have a very similar overall WER and BLEU score, system a has the highest WER of all six ASR models.
Our best model has almost half the word error rate and a 15 points higher BLEU score than the best commercial system.
The performance of all systems on a dialect level is similar to the overall performance.
The Innerschweiz dialect region is the easiest dialect, whereas the Wallis dialect is the most difficult one. Especially System b and d have a hard time with this dialect.
This is surprising as our model is trained with the most training data for the Bern and Zürich dialect region.
We have 10 and 45 times more training data for Zürich and Bern dialect regions than Innerschweiz.
 
 We see that unsupervised learning on 436K hours improves the scores when comparing our XLS-R-based model to the transformer baseline model.
 The differences across the various dialects for both models behave very similarly.
 We conclude that the reason for the different performance levels lies in the (finetuning) training and test data and does not stem from the transfer learning when finetuning a large XLS-R model to Swiss German.
 
\begin{table*}[!t]
\resizebox{\columnwidth}{!}{%
    \centering
    \begin{tabular}{l|cc|cc|cc|cc|cc|cc|cc|cc}
        \toprule 
        \multirow{2}{*}{\textbf{System}} & \multicolumn{2}{c}{\textbf{Overall}} & \multicolumn{2}{c}{\textbf{Basel}} & \multicolumn{2}{c}{\textbf{Bern}} & \multicolumn{2}{c}{\textbf{Graubünden}} & \multicolumn{2}{c}{\textbf{Innerschweiz}} & \multicolumn{2}{c}{\textbf{Ostschweiz}} & \multicolumn{2}{c}{\textbf{Wallis}} & \multicolumn{2}{c}{\textbf{Zürich}} \\
         & \textbf{WER} & \textbf{BLEU} &  \textbf{WER} & \textbf{BLEU} & \textbf{WER} & \textbf{BLEU} & \textbf{WER} & \textbf{BLEU} & \textbf{WER} & \textbf{BLEU} & \textbf{WER} & \textbf{BLEU} & \textbf{WER} & \textbf{BLEU} & \textbf{WER} & \textbf{BLEU} \\ 
         \midrule
        System a & $30.15$\% & $0.545$ & $29.99$\% & $0.540$ & $34.83$\% & $0.498$ & $28.72$\% & $0.561$ & $26.92$\% & $0.583$ & $31.33$\% & $0.540$ & $31.56$\% & $0.514$ & $27.70$\% & $0.575$ \\
        System b & $27.91$\% & $0.542$ & $29.47$\% & $0.520$ & $28.75$\% & $0.539$ & $25.79$\% & $0.555$ & $24.52$\% & $0.583$ & $28.10$\% & $0.545$ & $34.38$\% & $0.462$ & $24.36$\% & $0.593$ \\
        System c & $27.26$\% & $0.568$ & $28.35$\% & $0.554$ & $31.49$\% & $0.526$ & $24.71$\% & $0.600$ & $24.21$\% & $0.603$ & $26.70$\% & $0.576$ & $30.42$\% & $0.524$ & $24.94$\% & $0.600$ \\
        System d & $27.23$\% & $0.558$ & $28.58$\% & $0.532$ & $28.72$\% & $0.557$ & $24.96$\% & $0.576$ & $23.63$\% & $0.600$ & $26.76$\% & $0.565$ & $32.76$\% & $0.491$ & $24.85$\% & $0.587$ \\
        FHNW XLS-R        & $15.32$\% & $0.722$ & $16.30$\% & $0.702$ & $15.74$\% & $0.719$ & $14.32$\% & $0.736$ & $13.26$\% & $0.753$ & $16.45$\% & $0.710$ & $17.75$\% & $0.684$ & $13.41$\% & $0.749$ \\
        FHNW Transformer & $19.19$\% & $0.682$ & $21.24$\% & $0.663$ & $20.96$\% & $0.654$ & $17.29$\% & $0.703$ & $16.37$\% & $0.722$ & $18.58$\% & $0.688$ & $22.64$\% & $0.636$ & $17.30$\% & $0.708$ \\
        \bottomrule
    \end{tabular}%
    }
    \caption{Overall and per dialect region performance on the STT4SG-350 test.}
    \label{tab:results_full}
\end{table*}

To assess the influence of sentences containing named entities such as organization names and locations, we calculate the BLEU score on sentences containing such named entities.
We use spaCy to detect named entities in our test set.
The STT4SG-350 test set contains $7'148$ sentences with named entities.
In Table \ref{tab:results_ner} we report the BLEU score on sentences containing named entities.
The reported averages are macro-average precisions, whereas the corpus level statistics are micro-average precisions. Therefore those averages can not be compared directly.
The overall performance of all systems drops by about 10 BLEU points.
Sentences containing person names are tricky, whereas organizations and locations seem easier.

We see similar characteristics of our XLS-R-based model and transformer-based model on sentences containing named entities.

\begin{table}[tb]
    \centering
    \begin{tabular}{lccccc}
         \toprule
         System & Overall & Organisation & Person & Location & Miscellaneous \\
         \midrule
         System a & $0.456$ & $0.433$ & $0.398$ & $0.489$ & $0.471$ \\
         System b & $0.426$ & $0.402$ & $0.371$ & $0.445$ & $0.434$\\
         System c & $0.453$ & $0.432$ & $0.389$ & $0.487$ & $0.469$ \\
         System d & $0.446$ & $0.423$ & $0.393$ & $0.476$ & $0.461$\\
         FHNW XLS-R        & $0.614$ & $0.657$ & $0.514$ & $0.657$ & $0.596$\\ 
         FHNW Transformer & $0.591$ & $0.626$ & $0.510$ & $0.652$ & $0.555$ \\
         \bottomrule
    \end{tabular}
    \caption{BLEU score of sentences with named entities in the STT4SG-350 test set. The first column gives the overall BLEU score regardless of the named entity type, the other columns for the different named entity types.}
    \label{tab:results_ner}
\end{table}

In Table \ref{tab:per_speaker_performance} we report the per speaker BLEU score statistics.
The reported averages are again macro-average precisions, therefore not directly comparable with the corpus level statistics.  
For each speaker, we calculate his average BLEU score. In the following columns, we report the mean, standard deviation, lowest and highest average BLEU scores over all speakers.
We see that our systems are quite sensitive to individual speakers, since the average BLEU score across the speakers has a standard deviation of 7.2 and 8.9 BLEU points respectively.
Systems a and c are the most stable ones across the speakers.
A manual inspection showed that speakers with a low BLEU score in our ASR system could have a higher BLEU score on other systems, even though we have an overall higher BLEU score.

\begin{table}[tb]
    \centering
    \begin{tabular}{lcccc}
        \toprule
         System & Avg. BLEU & std & min & max \\
         \midrule
         System a & 0.483 & 0.023 & 0.441 & 0.547 \\
         System b & 0.475 & 0.038 & 0.389 & 0.571 \\
         System c & 0.484 & 0.021 & 0.420 & 0.530 \\
         System d & 0.478 & 0.032 & 0.422 & 0.571 \\
         FHNW XLS-R         & 0.657 & 0.072 & 0.497 & 0.785 \\
         FHNW Transformer  & 0.614 & 0.089 & 0.403 & 0.755 \\
         \bottomrule
    \end{tabular}
    \caption{Per speaker macro-average BLEU score statistics on the STT4SG-350 test set. For each speaker we calculate the average BLEU score and then the average, standard deviation, minimum and maximum over all speakers.}
    \label{tab:per_speaker_performance}
\end{table}

In Table \ref{tab:per_gender_performance} we report the macro-average BLEU score statistics per gender in the same way as for the speaker statistics. 
The performance on male and female voices does not depend on the gender for System a and c.
In case of the other systems we see a 1 BLEU point difference between male and female voices which is negligible.

\begin{table}[tb]
    \centering
    \begin{tabular}{lcc}
        \toprule
         System & male BLEU & female BLEU \\
         \midrule
         System a & 0.484 & 0.485  \\
         System b & 0.483 & 0.468   \\
         System c & 0.485 & 0.484 \\
         System d & 0.471 & 0.481 \\
         FHNW XLS-R        & 0.661 & 0.659 \\
         FHNW Transformer & 0.614 & 0.622 \\
         \bottomrule
    \end{tabular}
    \caption{Per gender macro-average BLEU scores on the STT4SG-350 test set.}
    \label{tab:per_gender_performance}
\end{table}

In Table \ref{tab:per_age_performance} we report the macro-average BLEU score statistics per age.
There is no clear trend visible in the results.
For the commercial systems c and d, the BLEU score is the highest on teen voices.
System a and our models reach the highest BLEU scores in the fifties age group.
System b, on the other hand, obtains its highest BLEU score for the thirties.
We conclude that the difficulties for an ASR system on the STT4SG-350 test set are evenly distributed across all age groups.

\begin{table}[tb]
    \centering
    \begin{tabular}{lcccccc}
        \hline
        \toprule
         System & teens & twenties & thirties & fourties & fifties & sixties \\
         \midrule
         System a & 0.482 & 0.486 & 0.479 & 0.477 & 0.509 & 0.489  \\
         System b & 0.441 & 0.464 & 0.496 & 0.469 & 0.471 & 0.489   \\
         System c & 0.514 & 0.483 & 0.492 & 0.479 & 0.469 & 0.486 \\
         System d & 0.510 & 0.470 & 0.490 & 0.471 & 0.487 & 0.475 \\
         FHNW XLS-R        & 0.685 & 0.654 & 0.679 & 0.638 & 0.700 & 0.664 \\
         FHNW Transformer & 0.641 & 0.601 & 0.659 & 0.601 & 0.665 & 0.617 \\
         \bottomrule
    \end{tabular}
    \caption{Per age group macro-average BLEU scores on the STT4SG-350 test set.}
    \label{tab:per_age_performance}
\end{table}

\subsection{Swiss Parliament Corpus}
The SPC corpus is not annotated with gender, age, and dialect meta data.
We limit the analysis of the SPC test set to the two best performing commercial system and the best FHNW model. In Table \ref{tab:results_spc} we display the overall BLEU score and WER for these systems.
On the SPC test set system d clearly outperforms system c by 4 BLEU points and a 3\% lower WER, whereas on the STT4SG-350 test set, system c achived a higher a 1\% BLEU score than system d.
Again, our XLS-R-based model has the lowest WER and highest BLEU score.

In Table \ref{tab:results_ner_spc} we repeat our named entity analysis on the SPC test set.
System d shows a significant higher BLEU score than system c on all named entity types.
Sentences containing person names are the most challenging ones leading to an overall lower BLEU score for all systems.

\begin{table}[tb]
    \centering
    \begin{tabular}{lcc}
         \toprule
         System & WER & BLEU \\
         \midrule
%         System a & n/a & n/a \\
%         System b & n/a & n/a \\
         System c & $36.46$\% & $0.460$ \\
         System d & $33.44$\% & $0.509$ \\
         FHNW XLS-R & $23.65$\% & $0.607$ \\ 
         \bottomrule
    \end{tabular}
    \caption{WER and BLEU score on the SPC test set.}
    \label{tab:results_spc}
\end{table}

\begin{table}[tb]
    \centering
    \begin{tabular}{lccccc}
         \toprule
         System & Overall & Organisation & Person & Location & Miscellaneous \\
         \midrule
%         System a & n/a & n/a & n/a & n/a & n/a \\
%         System b & n/a & n/a & n/a & n/a & n/a \\
         System c & $0.416$ & $0.427$ & $0.337$ & $0.405$ & $0.440$ \\
         System d & $0.476$ & $0.491$ & $0.415$ & $0.468$ & $0.485$\\
         FHNW XLS-R & $0.573$ & $0.607$ & $0.469$ & $0.559$ & $0.577$\\ 
         \bottomrule
    \end{tabular}
    \caption{BLEU score of sentences with named entities in the SPC test set. The first column gives the overall BLEU score regardless of the named entity type, the other columns for the different named entity types.}
    \label{tab:results_ner_spc}
\end{table}

%\clearpage

\subsection{Transcript Examples}

In Table \ref{tab:problematic_sentences} we list three examples of transcripts produced by the different ASR systems.
We list the German sentence in the column ground truth.
For each sentence we have seven recordings in the STT4SG-350 test corpus, one for each dialect region.
We show two or three transcripts produced by each system.

The first example contains an abbreviated name (A. M.).
Some of the recruited speakers voiced the abbreviation's punctuation while others did not.
Systems a-d correctly transcribed the punctuation but will be punished by the BLEU and WER calculation as they are not part of the ground truth.
In that specific example system, a-d created more accurate transcripts than our model.
All systems fail to transcribe A. M., when the speaker did not voice the punctuation.

The second example contains the English saying "last but not least", which is also common in Switzerland.
Systems a and c can create the correct English transcript. Our model produced the correct transcript in one case, and systems b and d failed on all utterances.

The third example contains a common named entity (FDP; a swiss political party also existing under the same name in Germany.) in a long sentence.
The first recording of this example is wrong. The speaker read "Reglementierung" instead of "Dereglementierung"; therefore, the transcript of all systems is perfect.
In the second recording, the speaker said "Dereglementierung" but all except our system failed to produce this antonym.
System a and c transcribed "die Reglementierung" instead and system b and d ignored it producing "Reglementierung".

\begin{table}[t]
    \centering
    \begin{tabular}{p{0.14\linewidth}|p{0.14\linewidth}|p{0.14\linewidth}|p{0.14\linewidth} | p{0.14\linewidth} | p{0.14\linewidth}}
        \toprule
        Ground Truth & System a & System b & System c & System d & FHNW XLS-R \\
        \midrule
        Darauf zeigte A.M. ihn an. & da drauf hat a punkt m punkt in angezeigt & darauf hat der a punkt n punkt ihn angezeigt & da daruf hat a punk m punkt in angezeigt & darauf hat der a punkt n punkt ihn angezeigt & darauf zeigte er an \\
        &&&&&\\
        & daraufhin hat der angezeigt & daraufhin hat der ami angezeigt & daraufhin hat der angezeigt & daraufhin hat der ami angezeigt & daraufhin zeigte ihn an \\
        \midrule
        Last but not least & last but not least & lasst not liest & last but not least & lasst not liest & last but not least \\
        & last but not least & la parmar liest & last but not least & la parmar liest & das bad ist \\
        & last but not least & last gleist & last but not least & last gleist & das bad noch die liste \\
        \midrule
        Für die Fraktion FDP wäre eine Dereglementierung wünschenswert und nicht umgekehrt. & für die fraktion fdp wäre die reglementierung wünschenswert und nicht umgekehrt & für die fraktion fdp wäre die reglementierung wünschenswert und nicht umgekehrt & für die fraktion fdp wäre die reglementierung wünschenswert und nicht umgekehrt & für die fraktion fdp wäre die reglementierung wünschenswert und nicht umgekehrt & für die fraktion fdp wäre diese reglementierung wünschenswert und nicht umgekehrt \\ % Falsch aufgenommen vom probanden
        &&&&&\\
        & für die fraktion fdp wäre eine die reglementierung wünschenswert und nicht umgekehrt & für die fraktion fdp wäre eine reglementierung wünschenswert und nicht umgekehrt & für die fraktion fdp wäre eine die reglementierung wünschenswert und nicht umgekehrt & für die fraktion fdp wäre eine reglementierung wünschenswert und nicht umgekehrt & für die fraktion fdp wäre eine dereglementierung wünschenswert und nicht umgekehrt \\
        \bottomrule
    \end{tabular}
    \caption{Transcription examples from the STT4SG-350 test set for five evaluated systems.}
    \label{tab:problematic_sentences}
\end{table}

\section{Conclusion}
The STT4SG-350 test set allows us to evaluate ASR systems on the seven most prominent Swiss dialects.
Some dialects are closer to standard German than others.
We would have estimated that ASR systems obtain higher scores on dialects closer to Standard German, but this is not the case.
For example, we obtain the highest BLEU score on the dialect region Innerschweiz with all evaluated ASR systems, except for system b.
This dialect region includes very strong dialects from rural regions.
It is more difficult to recruit speakers from rural regions than townspeople, which often speak a less pronounced dialect.
We assume that we failed to obtain enough speech data from rural regions to perform a more meaningful evaluation, even though we collected a large test corpus with the same amount of speech data for all seven dialect regions.
This will be assessed in-depth in future work.

Our system performs significantly better than the commercial ones on both test sets.
We evaluate all systems on two domains: parliament speeches and read-out news sentences.
System d obtained higher BLEU scores on parliament speech data than system c and performed on pair on read-out news data.
The characteristics of the two test corpora are similar in that both contain sentence-level recordings with a limited amount of background noise.

This analysis shows that our model trained on publicly available data can outperform all other systems significantly in this particular setting.
It is left to future work to evaluate the systems on a more general ASR setting containing free speech, longer recordings, dialogues, and more background noise.

\clearpage

\bibliographystyle{unsrtnat}
\bibliography{references}  %%% Uncomment this line and comment out the ``thebibliography'' section below to use the external .bib file (using bibtex) .

\end{document}